\documentclass[11pt,letterpaper]{article}
\usepackage{emnlp2017}
\usepackage{times}
\usepackage{latexsym}
\usepackage{booktabs}
\usepackage{amsmath}
\usepackage{graphicx}
\emnlpfinalcopy

\def\emnlppaperid{***}

\title{Lexical Bias In Essay Level Prediction}

\author{Georgios Balikas \\
Kelkoo Group, Grenoble \\
  {\tt geompalik@hotmail.com}}

\date{}

\begin{document}

\maketitle

\begin{abstract}
Automatically predicting the level of non-native English speakers given their written essays is an interesting machine learning problem. In this work I present the system \texttt{balikasg} that achieved the state-of-the-art performance in the CAp 2018 data science challenge among 14 systems. I detail the feature extraction, feature engineering and model selection steps and I evaluate how these decisions impact the system's performance. The paper concludes with remarks for future work. 
\end{abstract}

\section{Introduction}
Automatically predicting the level of English of non-native speakers from their written text is an interesting text mining task. Systems that perform well in the task can be useful components for online, second-language learning platforms as well as for organisations that tutor students for this purpose. In this paper I present the system \texttt{balikasg} that achieved the state-of-the-art performance in the CAp 2018 data science challenge among 14 systems.\footnote{url{http://cap2018.litislab.fr/competition-en.html}} 
In order to achieve the best performance in the challenge, I decided to use a variety of features that describe an essay's readability and syntactic complexity as well as its content. 
For the prediction step, I found Gradient Boosted Trees, whose efficiency is proven in several data science challenges, to be the most efficient across a variety of classifiers. 

The rest of the paper is organized as follows: 
in Section 2 I frame the problem of language level as an ordinal classification problem and describe the available data. 
Section 3 presents the feature extaction and engineering techniques used. 
Section 4 describes the machine learning algorithms for prediction as well as the achieved results. 
Finally, Section 5 concludes with discussion and avenues for future research.

\section{Problem Definition}
In order to approach the language-level prediction task as a supervised classification problem, I frame  it as an ordinal classification problem. In particular, given a written essay $x$ from a candidate, the goal is to associate the essay with the level $\ell\in\mathcal{L}$ of English according to the Common European Framework of Reference for languages (CEFR) system. Under  CEFR there are six language levels $\ell$, such that $\mathcal{L}= \{A_1, A_2, B_1, B_2, C_1, C_2\}$.  In this notation, $A_1$ is the beginner level while $C_2$ is the most advanced level. Notice that the levels of $\mathcal{L}$ are ordered, thus defining an ordered classification problem. In this sense, care must be taken both during the phase of model selection and during the phase of evaluation. In the latter, predicting a class far from the true should incur a higher penalty. In other words, given a $C_1$ essay, predicting $A_1$ is worse than predicting $B_2$, and this difference must be captured by the evaluation metrics. 

In order to capture this explicit ordering of $\mathcal{L}$, the organisers proposed a cost measure that uses the confusion matrix of the prediction and prior knowledge in order to evaluate the performance of the system. In particular, the meaures uses writes as:
\begin{equation}
 E = \frac{100}{n}\sum\limits_{i=1}^{6}\sum\limits_{j=1}^{6}C_{i,j}N_{i,j} \label{eq:error}
\end{equation}
where $C$ is a cost matrix that uses prior knowledge to calculate the misclassification errors and $N_{i,j}$ is the number of observations of class $i$ classified with category $j$. The cost matrix $C$  is given in Table \ref{tbl:cost}. Notice that, as expected, moving away from the diagonal (correct classification) the misclassification costs are higher. The biggest error (44) occurs when a $C_2$ essay is classified as $A_1$. On the contrary, the classification error is lower (6) when the opposite happens and an $A_1$ essay is classified as $C_2$. Since $C$ is not symmetric and the costs of the lower diagonal are higher, the penalties for misclassification are worse when essays of upper languages levels (e.g., $C_1, C_2$) are classified as essays of lower levels.

\begin{table}\small\centering
 \begin{tabular}{lcccccc}\toprule 
 
 &$A_1$ & $A_2$ & $B_1$ & $B_2$ & $C_1$ & $C_2$  \\\midrule
$A_1$& 0 & 1& 2&     3&      4&      6\\
$A_2$& 1 & 0& 1&      4&     5&     8\\
$B_1$& 3 & 2& 0&     3&      5&      8 \\
$B_2$& 10& 7& 5&   0&     2&      7\\
$C_1$& 20& 16&12&     4&    0&     8\\
$C_2$& 44& 38&32&      19&      13&     0\\\bottomrule
 \end{tabular}
\caption{Cost matrix used to calculate the miscalssification error described in Eq. \eqref{eq:error}. }\label{tbl:cost}
\end{table}

\paragraph{Dataset} The data used in this work were released in the framework of the CAp 2018 competition and are an excerpt of \cite{geertzen2013automatic,huang2018dependency}. The competition's goal was to evaluate automated systems that performed well on the task. The organisers released two datasets: the training and the test data. The test data became available only a few days before the competition end without the associated language level and were used only for evaluation. The evaluation was performed by the organisers after submitting the system prediction in a text file as frequently done in such competitions. 

\begin{figure}\centering
 \includegraphics[scale=0.32]{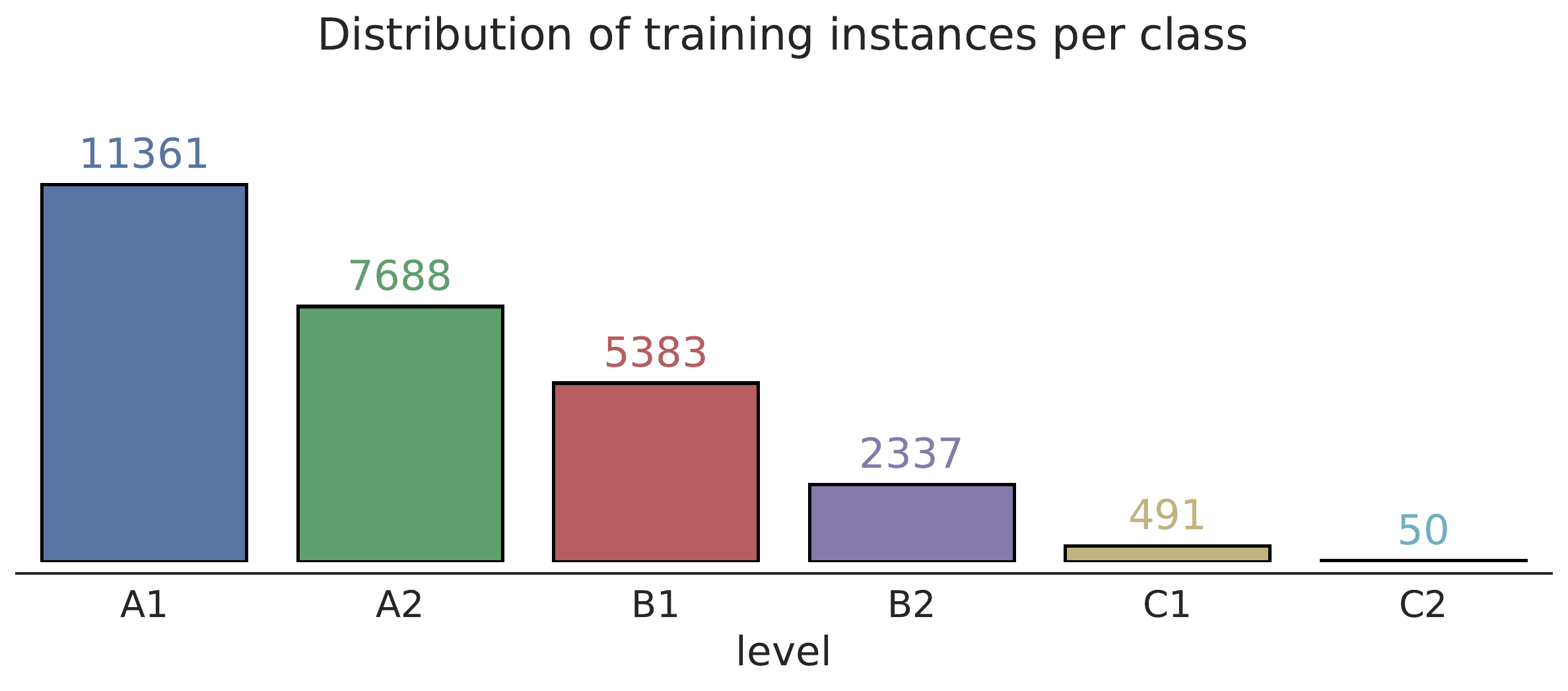}
 \caption{The distribution of essays according to the CERF levels in the training data.}\label{fig:training_distribution}
\end{figure}

The training data consist of 27,310 essays while the test data contain 13,656 essays. Figure \ref{fig:training_distribution} illustrates the distribution of essays according to their level. The classification problem is unbalanced as there are far more training examples for the first levels (e.g., $A_1$) than for the rest. The organisers announced that they performed a stratified split with respect to the level label, so we expect similar distributions for the test data.\footnote{At the time of writing of this paper, the test data have not become publicly available.} 

The released data consist of the essay text as well as various numerical features. The numerical  features  are either statistics calculated on the essay text (length, number of sentences/syllables etc.) or indexes that try to captrue the readability and complexity of the essays, like the Coleman and Flesch families of indexes. Table \ref{tbl:essay_stats} presents basic statistics that describe the essay text. 

\begin{table}\small\centering
 \begin{tabular}{lc}\toprule
 Description & Value \\\midrule
 Number of essays & 13,656 \\
 Vocabulary size & 38,337 \\
 Avg. essay length (words) & 80.22 \\ 
 Avg. essay length (sentences) & 6.75 \\ 
 \bottomrule
 \end{tabular}
\caption{Basic statistics for the released essays.}\label{tbl:essay_stats}
\end{table}

\section{Feature Extaction}
In this section I present the extracted features partitioned in six groups and detail each of them separately. 

\paragraph{Numerical features} Most of the features in this family were provided by the challenge organisers using the v0.10-2 of the R koRpus package.\footnote{\url{https://cran.r-project.org/web/packages/koRpus/vignettes/koRpus_vignette.pdf}} For a full list of these features, please visit the challenge website.\footnote{\url{http://cap2018.litislab.fr/competition_annexes_EN.pdf}} During the preliminary exploratory analysis I found some of the features released by the organisers to be inaccurate. Hence, I recalculated the number of sentences, words, letters per essay and I added the Gunning Fog index, which estimates the number of years of formal education a person needs to understand an English text on the first reading using the  Python \texttt{textstat} package.\footnote{https://github.com/shivam5992/textstat} Also, I added the number of difficult words in an essay using the lists of difficult words of \texttt{textstat}, the number of mispelled words using the GNU dictionary\footnote{\url{ftp://ftp.gnu.org/gnu/aspell/dict/0index.html}}, the number of duplicate words in the essay as well as the number of words with inverse document frequency (idf) smaller then the average idf of the corpus words. Overall, there are 66 numerical features in this family. 

\paragraph{Language models} For each essay I calculated its probability under two language models: one trained with the essays belonging to the labels $\ell\in\{A_1, A_2, B_1\}$ and another trained on the essays of $\ell\in\{B_2, C_1, C_2\}$. For this purpose, I used trigram language models with modified Kneser Ney smoothing, with $n$-grams of a lower order ($n\in\{1,2\}$) as a back-off mechanism \cite{heafield2013scalable} using the implementation of \cite{heafield2011kenlm}. As language models can easily overfit when trained on small corpora, I decided to replace words with less than 10 occurrences by their part-of-speech (POS) tags and numbers by the special token ``$<number>$''. The hope is that language models will capture the more complicated patterns we expect higher-level users to use. The free parameters concerning the use of language models like the optimal value of $n\in{1,2,3}$, the decision whether to ignore or replace low frequency words with their POS etc. were made using stratified 3-fold cross-validation in the training data. The same applies for every major decision in the feature extraction process, unless otherwise stated.    

\paragraph{Word Clusters} Word embeddings are dense word vectors that have been show to capture the semantics of words. I represented a given essay using word clusters \cite{balikas2017effectiveness} calculated by applying $k$-Means ($k=1,000$) on the ConceptNet embeddings \cite{speer2017conceptnet}. Clustering the words of the corpus vocabulary generates semantically coherent clusters. In turn, to represent a document I used a binary one-hot-encoded representation of the clusters where the essay words belong to. For instance, in our case where  each of the vocabulary words belongs to one of  1,000 clusters, an essay is represented by a 1,000-dimensional binary vector. The non-zero elements of this vector are the ids of clusters where the essay words belong to.  

\paragraph{Topic Models} Topic models are a class of unsupervised models. They are generative models in that they define a mechanism on how a corpus of documents is generated. In this work I used Latent Dirichlet Allocation (LDA) \cite{blei2003latent} in order to obtain dense document representations that describe the topics that appear in each document. During the inference process, these topics, that are multinomial distributions over the corpus vocabulary, are uncovered and each document is represented by a mixture of them. I used a custom Python LDA implementation\footnote{https://github.com/balikasg/topicModelling} and concatenated the per-document topic distributions obtained when training LDA with 30, 40, 50 and 60 topics.  For each number of topics, I ran the inference process  for 200 burn-in iterations, so that the collapsed Gibbs sampler converges, and then sampled the topic distributions every 10 iterations until 50 in order to obtain uncorrelated samples. 

\paragraph{Part-of-Speech tags} POS tags are informative text representations that can capture the complexity of the expressions used by an author. Intuitively, beginners use less adjectives and adverbs for example compared to more advanced users of a language.  To capture this, I obtained the sequence of POS tags of an essay using SpaCy.\footnote{https://spacy.io/}  Then to represent the POS sequences I used $n$-grams ($n=1$) and encoded them as bag-of-words. 

\paragraph{Essay text}
Last, I explicitly encoded the content of the essay using its bigram bag-of-words representation. In order to limit the effect of frequent terms like stopwords I applied the idf weighting scheme. 

\section{Model Selection and Evaluation}
As the class distribution in the training data is not balanced, I have used stratified cross-validation for validation purposes and for hyper-parameter selection. 
As a classification1 algorithm, I have used gradient boosted trees trained with gradient-based one-side sampling as implemented in the Light Gradient Boosting Machine toolkit released by Microsoft.\footnote{\url{https://github.com/Microsoft/LightGBM}}. The depth of the trees was set to 3, the learning rate to 0.06 and the number of trees to 4,000. Also, to combat the class imbalance in the training labels I assigned class weights at each class so that errors in the frequent classes incur less penalties than error in the infrequent.

\begin{figure}\centering
 \includegraphics[scale=0.26]{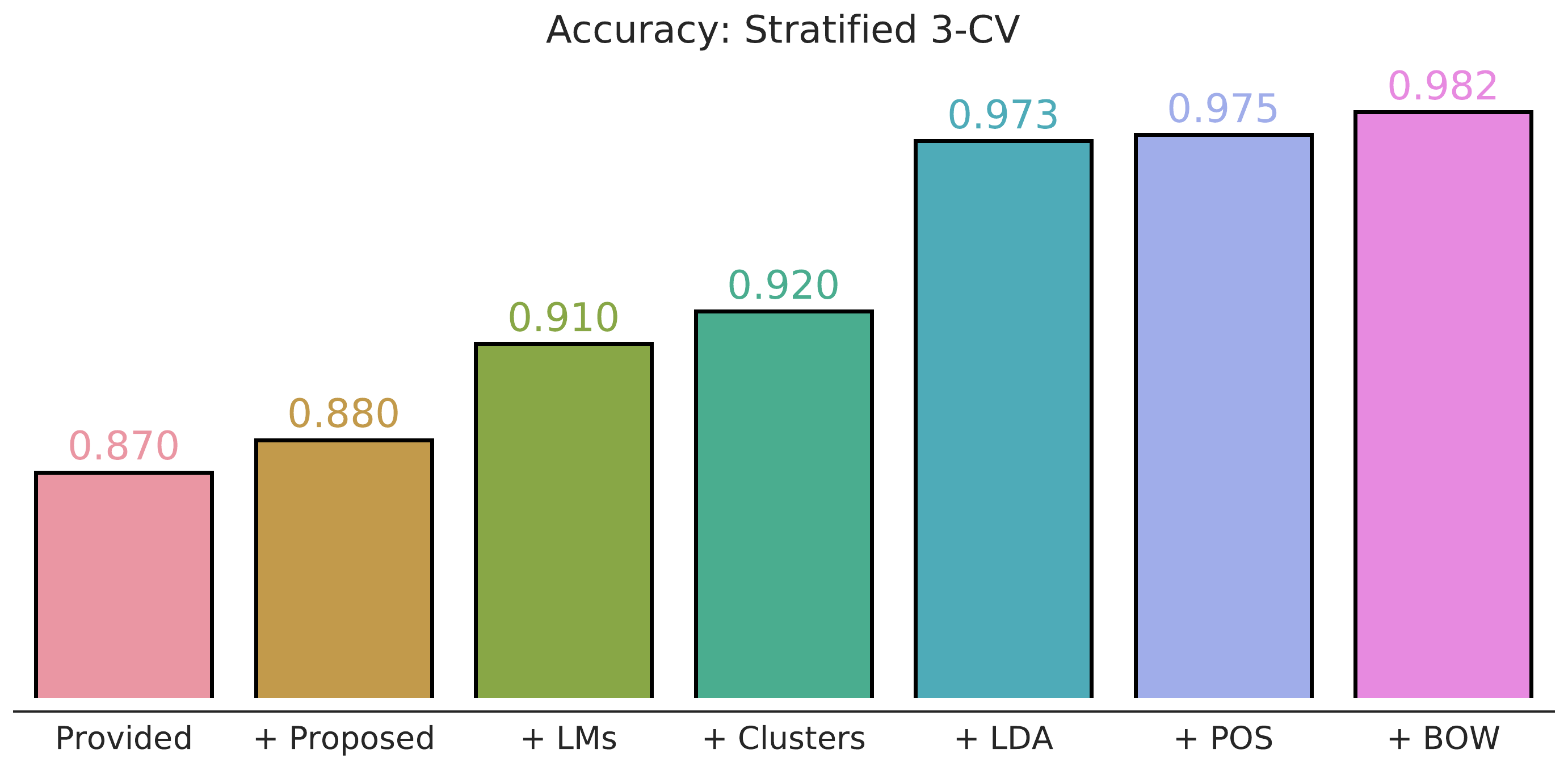}
 \caption{The accuracy scores of each feature set using 3-fold cross validation on the training data.}\label{fig:perf}
\end{figure}

\paragraph{Evaluation}
Figure \ref{fig:perf} illustrates the performance the Gradient Boosted Trees achieve on each of the feature sets. Complementary and for reference, Table \ref{tbl:acc_scores} presents the accuracy scores of each feature set. Notice that the best performance is obtained when all features are used. Adding the per-document topic distributions infered by the topic models seems to improve the results considerably.  

\begin{table}\small\centering
 \begin{tabular}{lc}\toprule
 Features & $E$ \\\midrule
 + Numerical Features & 23.52 \\
 + Language Models  &  14.00\\
 + Clusters & 14.20 \\
 + Latent Dirichlet Allocation &  5.74\\ 
 + Part-Of-Speech tags&  5.52 \\ 
 + Bag-of-words & 4.97 \\
 \bottomrule
 \end{tabular}
\caption{Stratified 3-fold cross-validation scores for the official measure of the challenge.}\label{tbl:acc_scores}
\end{table}

In order to better evaluate the effect of each family of features without the bias of the ordering of adding families of features, Table \ref{tbl:ablation_study} presents an ablation study. The Table presents the scores achieved when using all features as well as the scores achieved when removing a particular family of features. In this sense, one can estimate the added value of each family. From the table we notice that the most effective features are the numerical features, the document distributions learned with LDA and the scores from the language models. When removing the features of this family we notice the biggest reduction in performance. For reference, the table presents the performance of a tuned Logistic Regression with the same class weights for the class imbalance. Gradient Boosted trees outperform Logistic Regression by a large margin in terms of $E$, proving their efficiency for classification tasks. 

\begin{table}\small\centering
 \begin{tabular}{lcc}\toprule 
 Features & Error ($E$) & Accuracy  \\\midrule
 All features (Winning solution)     & 4.97  & 98.2 \\
 All features (Log. Regression) & 10.10 & 97.2 \\
 \midrule
 - Numerical Features          & 14.65 & 95.6 \\
 - Language Models             & 5.66  & 98.1 \\
 - Clusters                    & 4.99  & 98.1\\
 - Latent Dirichlet Allocation & 7.14  & 97.3 \\ 
 - Part-Of-Speech tags         & 5.01  & 98.1\\ 
 - Bag-of-words                & 5.52  & 97.7\\
 \midrule
 Reduced feature set           & 4.90  & 98.2 \\
 \bottomrule
 \end{tabular}
\caption{Ablation study to explore the importance of different feature families.}\label{tbl:ablation_study}
\end{table}

Another interesting point concerns the effect of some features on the two evaluation measures that Table \ref{tbl:ablation_study} presents. Notice, for instance, that while language models are quite important for the custom error metric $E$ (Eq. \ref{eq:error}) of the challenge, their effect is smaller for accuracy. This suggests that adding them reduces the size of the errors considerably, but does not increase much the correctly classified instances. 

The last observation on the impact of the features on the evaluation measures motivates an error analysis step to examine the errors the model produces. Table \ref{tbl:confusion_matrix} presents the confusion matrix of the 3-fold cross-validated predictions in the training data. As expected, the majority of examples appear in the diagonal denoting correct classification. Most of the errors that occur are in neighboring categories suggesting that it can be difficult to differentiate between them. Lastly,    very few misclassification errors occur between categories that are far with respect to the given order of language levels which suggests that the system successfully differentiates between them.

\begin{table}\small\centering
 \begin{tabular}{lcccccc}\toprule 
 
 &$A_1$ & $A_2$ & $B_1$ & $B_2$ & $C_1$ & $C_2$  \\\midrule
$A_1$& 11,224& 54&      3&      0&      1&      0\\
$A_2$& 99&   7,531&     42&      0&      0&      0\\
$B_1$& 30&     95&   5,297&     23&      7&      1 \\
$B_2$& 0&      4&     32&   2,273&     14&      1\\
$C_1$& 7&      2&      7&     35&    465&     19\\
$C_2$& 1&      2&      2&      6&      4&     29\\\bottomrule
 \end{tabular}
\caption{Confusion matrix of the 3-fold stratified cross validation. The $C_{i,j}$ value is the number of predictions known to be in group $i$ and predicted to be in group $j$. Notice how most of the mis-classification errors occur between close categories. }\label{tbl:confusion_matrix}
\end{table}

\section{Conclusion} In this work I presented the feature extraction, feature engineering and model evaluation steps I followed while developing \texttt{balikasg} for CAp 2018 that was ranked first among 14 other systems. I evaluated the efficiency of the different feature groups and found that readbility and complexity scores as well as topic models to be effective predictors. Further, I evaluated the the effectiveness of different classification algorithms and found that Gradient Boosted Trees outperform the rest of the models in this problem. 

While in terms of accuracy the system performed excellent achieving 98.2\% in the test data, the question raised is whether there are any types of biases in the process. For instance, topic distributions learned with LDA were valuable features. One, however, needs to deeply investigate whether this is due to the expressiveness and modeling power of LDA or an artifact of the dataset used. In the latter case, given that the candidates are asked to write an essay given a subject \cite{geertzen2013automatic} that depends on their level, the hypothesis that needs be studied is whether LDA was just a clever way to model this information leak in the given data or not. I believe that further analysis and validation can answer this question if the topics of the essays are released so that validation splits can be done on the basis of these topics.

\section*{Acknoledgements} I would like to thank the organisers of the challenge and NVidia for sponsoring the prize of the challenge. The views expressed in this paper belong solely to the author, and not necessarily to the author's employer.

\bibliography{emnlp2017}
\bibliographystyle{emnlp_natbib}

\end{document}